\documentclass[12pt]{article}

\usepackage{scicite}
\usepackage{times}
\usepackage{amsmath,mathrsfs}
\usepackage{float}
\usepackage{amsfonts}
\usepackage{graphicx}
\usepackage{epstopdf}
\usepackage{caption}
\usepackage{listings}
\usepackage{url}
\usepackage{bm}
\usepackage{subcaption}
\usepackage{soul}
\usepackage[shortlabels]{enumitem}
\usepackage{soul}

\newcommand{\ignore}[1]{}
\newcommand{\out}[1]{}
\newcommand{\OUT}[1]{}

\makeatletter
\DeclareUrlCommand\ULurl@@{%
  \def\UrlLeft{\uline\bgroup}%
  \def\UrlRight{\egroup}}
\def\ULurl@#1{\hyper@linkurl{\ULurl@@{#1}}{#1}}
\DeclareRobustCommand*\ULurl{\hyper@normalise\ULurl@}
\makeatother

\newenvironment{sciabstract}{%
\begin{quote} \bf}
{\end{quote}}

\title{Metal Blooming: Creating Complex Metal Structures Using Laser-Forming Origami} 
\title{Metal Blooming: Making Complex Metal Structures Using Laser-Forming Origami is Possible} 
\title{Metal Blooming: Making Complex Metal Structures Using Laser-Forming Origami by Imitating Flower Blooming} 
\title{Metal Blooming: Manufacturing Complex Metal Structures Using Laser-Forming Origami that Resembles Flower Blooming} 
\title{Metal Blooming: Manufacturing Freeform Metal Structures Via Laser-Forming Origami Mimicking Flower Blooming} 
\title{Metal Blooming: Freeform Metal Structures Manufactured via Laser-Forming Origami Imitating Flower Blooming} 
\title{Metal Blossom: Laser Forming Complex and Freeform Metal Structures Imitating Flower Blooming}

\author
{Yue Hao$^{1}$, Peiwen J. Ma$^{2}$, Huaishu Peng$^{3}$, \\Edwin A. Peraza Hernandez$^{2}$, Jyh-Ming Lien$^{1\ast}$\\
\\
\normalsize{$^{1}$Department of Computer Science, George Mason University, Fairfax, VA 22030}\\
\normalsize{$^{2}$Department of Mechanical and Aerospace Engineering, University of California, Irvine,} \\ \normalsize{Irvine, CA 92697}\\
\normalsize{$^{3}$Department of Computer Science, University of Maryland, College Park, MD 20742}\\
\\
\normalsize{$^{\ast}$To whom correspondence should be addressed; E-mail: \url{jmlien@gmu.edu}.}
}

\date{}

\usepackage{xcolor} 

\usepackage{graphicx}
\usepackage[normalem]{ulem}
\usepackage[hidelinks]{hyperref}
\usepackage{soul}

\usepackage{graphicx}
\begin{document} 

\baselineskip22pt

\maketitle 

\begin{sciabstract}

{\begin{sloppypar}
For centuries, human civilizations devised metal forming techniques 
to make tools and items; yet, customized metal forming remains costly and intricate.
\emph{Laser-forming origami} (\emph{lasergami}) is a metal forming process where a laser beam \emph{cuts} and \emph{folds} a planar metal sheet to form a three-dimensional (3D) shape. 
Designing foldable structures formable by lasers, 
however, has long been a trial-and-error practice that requires significant mental effort and hinders the possibility of creating practical structures.  
This work demonstrates for the first time that lasergami can form a freeform set of metallic structures previously believed to have been impossible to be laser-formed.
This technological breakthrough is enabled by new computational origami methods that imitate flower blooming 
and optimize laser folding instructions. 
Combined with new ideas that address laser line of sight 
and minimize fabrication energy, we report a low-cost manufacturing framework that can be readily adopted by hobbyists and professionals alike.
\end{sloppypar}
}

\end{sciabstract}




%
%

\begin{figure}[]
    \centering
      {\includegraphics[width=1\textwidth]{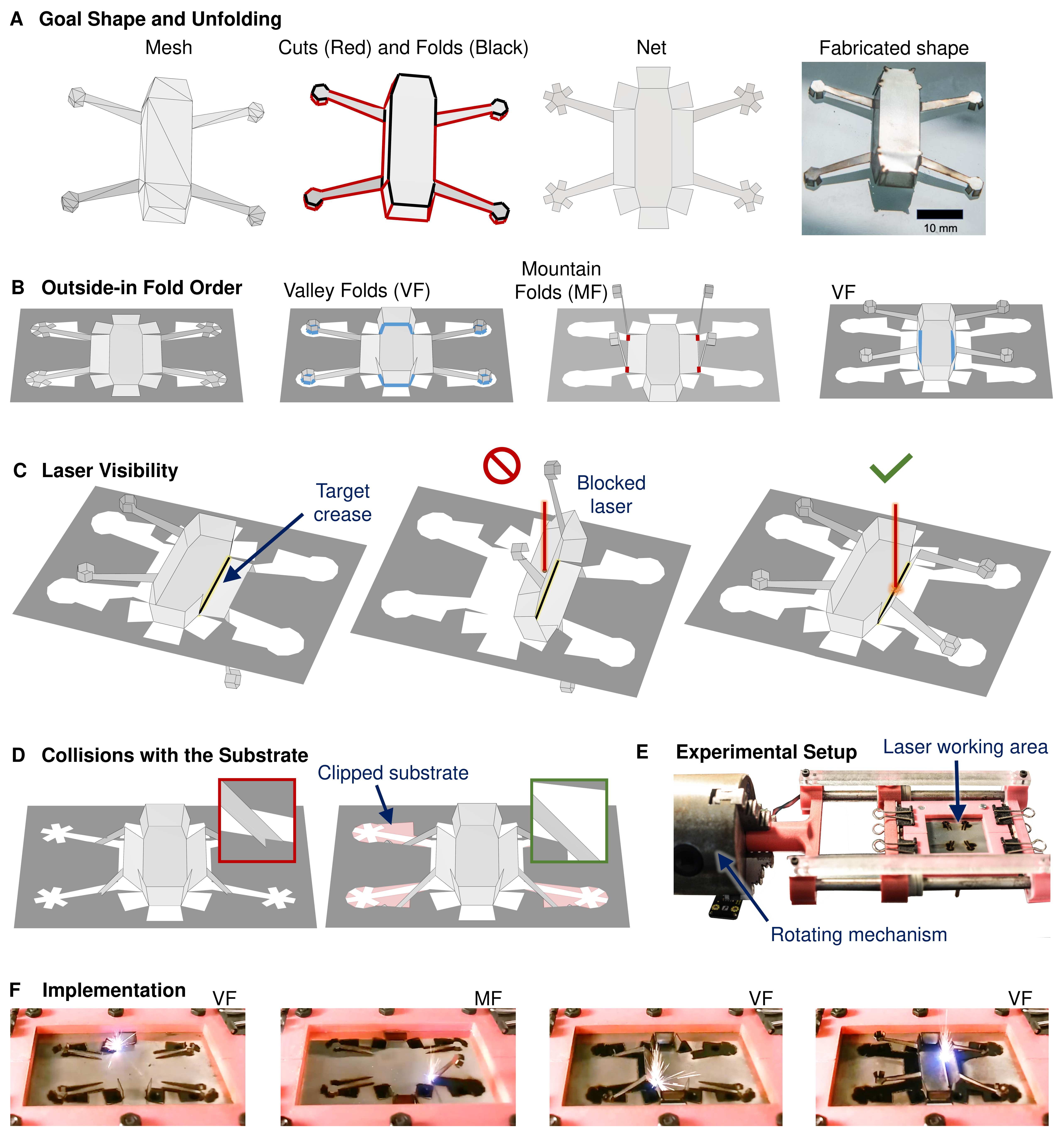}}%

    \caption{
    \textbf{Computational design and fabrication of a metallic drone frame using lasergami}.
    \textbf{(A)} 
    Fabricated drone frame next to the associated goal mesh, cut and fold edges, and net.
    \textbf{(B)}
    Edges are folded according to their distance (number of folds) from the stationary face. 
    \textbf{(C)} 
    The laser line of sight should not be blocked for any crease.
    \textbf{(D)} 
    Faces of the sample should not collide with the substrate during folding. 
    \textbf{(E)} 
    Rotary placement with respect to the laser working area. 
    \textbf{(F)} 
    Chronological series showing fabrication of the drone frame (Movie~S1). 
    }
    \label{fig:overview}
\end{figure}

\begin{figure*}[ht!]
    \centering
        \makebox[\textwidth][c]{\includegraphics[width=1\textwidth]{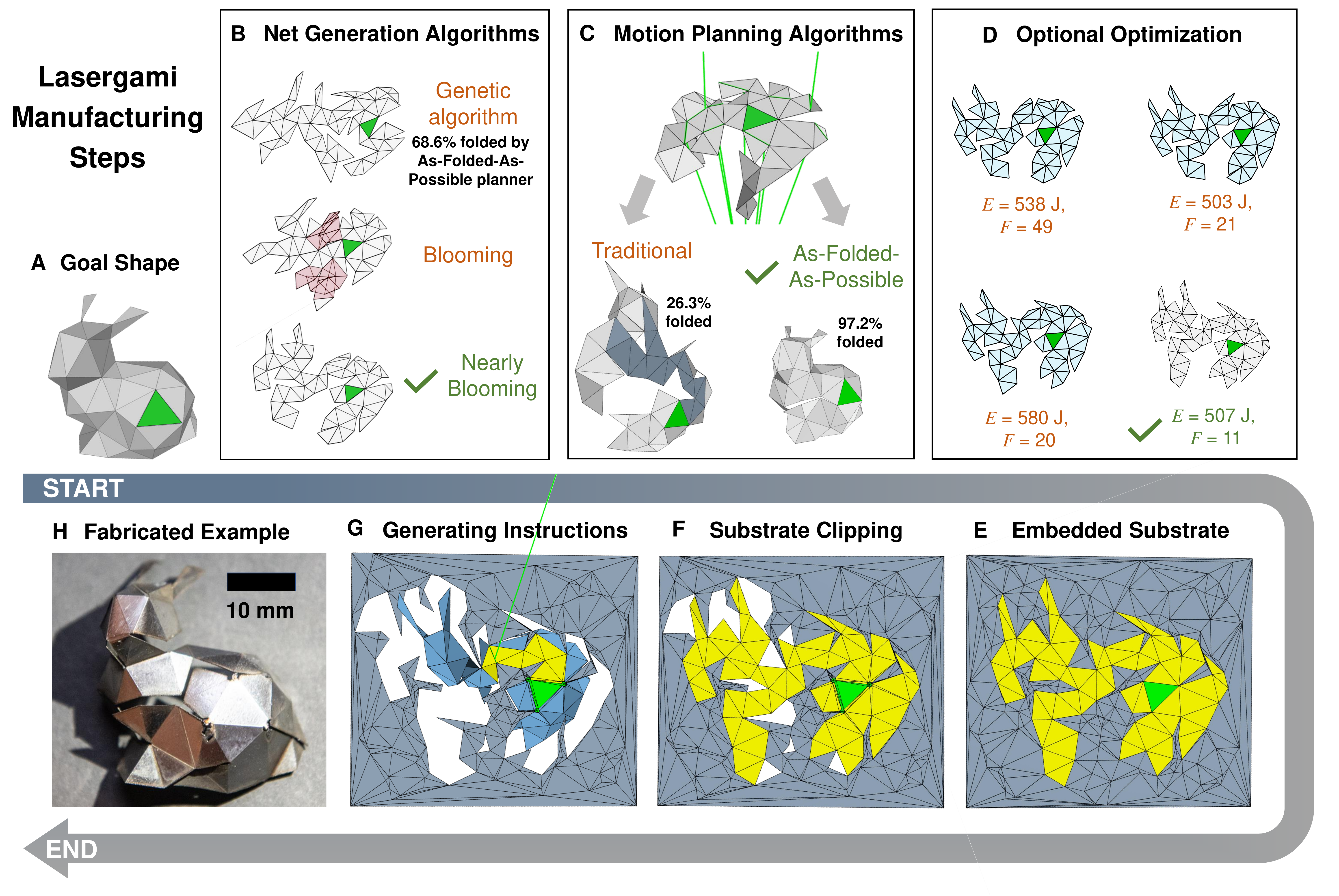}}%
    \caption{
    \textbf{
    Computational design of lasergami.} 
    \textbf{(A)} User input. 
    \textbf{(B)}~The NB algorithm unfolds the input onto a net that is significantly more foldable than those created by the other approaches. 
    \textbf{(C)} The FP algorithm folds the net significantly closer to the goal shape than the traditional motion planner. 
    \textbf{(D)} Optional optimization makes further improvements to the net. The fabrication energy $E$ and the number of clipped substrate faces $F$ are considered as minimization objectives (see supplemental materials for the formulation of $E$). 
    \textbf{(E)} Mesh incorporating the substrate with the net. 
    \textbf{(F)}~Clipping faces to eliminate substrate intersections.  
    \textbf{(G)}~Generating instructions for the laser manipulator.
    \textbf{(H)} Laser-formed bunny. 
    }
    \label{fig:pipeline}
\end{figure*}

\begin{figure}
    \centering
        \includegraphics[width=0.9\textwidth]{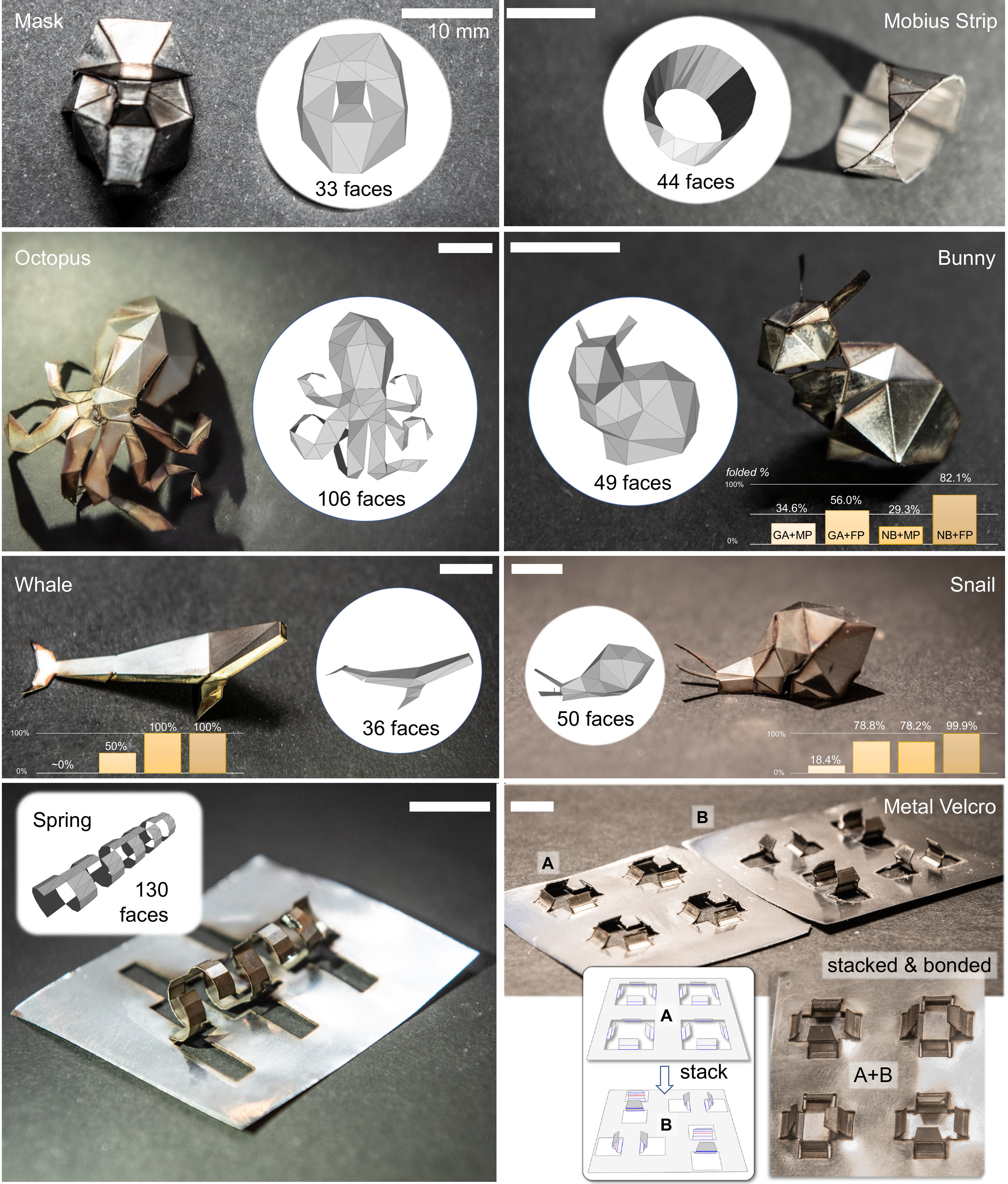}
    \caption{\textbf{Lasergami structures}. Low-cost laser forming with computational folding instructions enabled flexible, rapid, and cost-effective fabrication of complex shapes and tools (Table~S2, Movies S2-S3), opening accessible routes for laser cutter users and hobbyists alike. 
    Despite the many constraints of laser forming,
    there is evidently extensive flexibility towards arbitrary shape creation and control through the use of our novel fabrication instructions and net design methods. 
    }
    \label{fig:results}
\end{figure}

Metallic structures have played critical roles in human history, ranging from the formation of mega-scale 
ship hulls to micro/nano-scale devices~\cite{reiser2020metals,liu2018nano}. 
Since ancient times, civilizations have employed metal forming processes such as die punching, 
rolling, and bending that involve manual handling and hard tooling to form geometries from iron, steel, or aluminum~\cite{frazier2014metal}. 
However, customized metal forming to date remains costly, time-consuming, and labor-intensive. 
Laser-forming origami uses a laser to repeatedly induce localized thermoelastic expansion on metallic surfaces and generate plastic deformations that emulate folding, forming 3D metallic structures~\cite{cheng2004process,kim20093d} without human intervention~\cite{lazarus2017laser}.
Laser-forming origami begins with the cutting of a \emph{net}~\cite{hernandez2019unfolding}, a planar surface that can be continuously folded to transform into a goal shape under fine-tuned conditions for the speed, power, and laser beam size
\cite{lazarus2019self}.
Compared with traditional mechanical forming processes, laser forming with negligible springback~\cite{dearden2003some} has seen use in fabricating sensitive electronics and as a forming tool for astronauts in zero-gravity environments~\cite{namba1986laser,pique2012laser}. 
Compared with prevailing additive manufacturing methods~\cite{6678531}, laser-forming origami is generally faster and less costly, and structures created with laser forming often have superior electrical conductivity and strength than those made with 3D metal printing~\cite{Lazarus2018OrigamiIR}.
Recent research has shown that laser-induced origami performed using budget laser cutters has potential in reducing fabrication costs of metallic~\cite{lazarus2017laser}, acrylic~\cite{mueller2013laser}, and polyethylene terephthalate 
structures~\cite{sargent2019heat}, increasing the accessibility of laser-induced origami to the general public.
The problem faced is that a typical laser cutter is believed to be incapable of creating practical structures beyond those with a few folds as demonstrated by recent works~\cite{Lazarus2018OrigamiIR,lazarus2019self}, due to its limited degrees of freedom. 
Indeed, this limitation is universal to many origami or kirigami-based fabrication routes that use external light or focused ion beams~\cite{liu2018nano}.
While folding is one of the fundamental tools used by Nature to transform a structure into its functional form \cite{dobson2003protein}, 
defining universal rules for designing a foldable structure from its goal shape has puzzled researchers for centuries \cite{o2008unfolding}. 
It is not surprising that designing laser-formable structures has remained a trial-and-error practice requiring significant mental effort and intensive labor, 
given that the designer must additionally consider laser-forming constraints such as laser line of sight and sample-substrate collisions shown in Fig.~\ref{fig:overview}. 
This work demonstrates for the first time, through computational origami and experimental implementation, that fabricating complex and practical metal structures from planar sheets via \emph{laser cutting and folding} (a process herein called \textit{lasergami}) is possible with its capability exemplified by the drone frame in Fig.~\ref{fig:overview}(\textbf{A}), the metallic bunny in Fig.~\ref{fig:pipeline}, and the diverse set of shapes in Fig.~\ref{fig:results}. 
Aiming at broadening the awareness and advances of lasergami, instructions for creating a lasergami vibrating robot are provided in Fig.~S17 and Movie~S4. 

The most used folding modes for lasergami are 
the temperature gradient mechanism (TGM) and the buckling mechanism (BM). 
In BM, the laser scans at a low speed  allowing heat to penetrate through the thickness of the metal sheet, causing upward or downward folding depending on the initial force loading condition~\cite{arnet1995extending,li2000buckling}. 
Because these initial conditions such as prebending should be applied on the entire metal sheet in a constant orientation, BM is somewhat limited to forming simple shapes.
In TGM, the laser rapidly scans across the crease location, and when the workpiece is cooled, the sheet folds upwards towards the laser source~\cite{shen2009modelling}.
This work uses TGM with the calibration settings provided in Table~S1, 
achieving bidirectional folding through the added rotary 
shown in Figs.~\ref{fig:overview}(\textbf{E}) and S3.
Fig.~S1 shows calibrated setups for sheets with different thicknesses.

For lasergami, the unique constraints shown in Fig.~\ref{fig:overview}(\textbf{B}-\textbf{D}) must be satisfied in addition to those from the classic polyhedral unfolding problem (Fig.~S4).
The \emph{outside-in folding order} shown in Fig.~\ref{fig:overview}(\textbf{B}), where the arms are folded before the central body of the drone frame, allows the calibration of the laser beam parameters (carried out on a sample having one face orthogonal to the laser beam (Fig.~S1)) to be applicable to the folding of every crease in this model. 
Laser-to-crease \emph{visibility} must also be ensured when folding each crease (Fig.~\ref{fig:overview}(\textbf{C})).
Unlike traditional polyhedral folding using paper in which the entire net is cut and then folded, lasergami
must iterate between cutting and folding to maximize fabrication reliability (Movie~S1). 
Consequently, it is necessary to consider potential collisions between the net and the substrate (Fig.~\ref{fig:overview}(\textbf{D}), Figs.~S2,~S6-S8)). 
We call a net of a 3D structure satisfying all these constraints \emph{laser-formable}. 

\paragraph{Methodology.}
Human or algorithmic designers of laser-formable structures
must make the decision on where (which edges of) the 3D goal shape, represented as a surface mesh, should be cut so the flattened 2D net is laser-formable. 
Consequently, our proposed computational framework adopts an iterative process of design and evaluation. 
Because a 3D surface mesh can be flattened onto many nets and a net can be folded in many ways, the evaluation step must plan and simulate the entire fabrication process. 
Compared with a self-folding net~\cite{hernandez2019unfolding}, lasergami encounters different challenges from its aforementioned special constraints. Combined with the inherent high dimensionality of the design space 
that inhibits analytical solutions, these differences make manual design nearly impossible. 


Two algorithms devised to overcome these challenges of lasergami are \textbf{nearly-blooming} (NB) net generation and \textbf{as-folded-as-possible} (FP) motion planning. 
We introduce these through the manufacturing flowchart in Fig.~\ref{fig:pipeline} and in the following sections, providing additional details in the supplementary materials. 
%
%
In Fig.~\ref{fig:pipeline}, the NB algorithm cuts the goal mesh along its edges followed by motion planning that finds the folding motion satisfying the remaining constraints. 
If such a folding motion path does not exist, the planner finds a path to bring the structure as close to the fully folded configuration as possible. 
Additional optimization (Fig.~\ref{fig:pipeline}\textbf{(D)}) then finds slightly altered nets that still satisfy all the lasergami constraints with overall improved performance. 
The substrate, which has its own mesh  (Fig.~\ref{fig:pipeline}\textbf{(E)}), has $F$ faces \emph{clipped} or \emph{folded} (Figs.~\ref{fig:pipeline}\textbf{(F)}, S14, S15) to ensure the net does not collide with the substrate (Alg.~S2). Finally, instructions for the laser beam (path, speed, power) are generated and passed to the laser cutter (Fig.~\ref{fig:pipeline}\textbf{(G)}).
Each step handles different but interdependent aspects of the manufacturing process;
thus, 
the influence of early steps on the performance of the next ones must be carefully assessed.
For example, 
we deliberately ensure that the NB unfolding can reduce chances of self-collisions during folding (Fig.~{S13}) and areas of clipped substrate  (Fig.~{S14}). 
Figure~\ref{fig:pipeline}\textbf{(B)} shows an example of a NB unfolding which, when compared with a net created from genetic algorithms (GA), can be laser folded closer to the goal 
(97.2\% vs. 68.6\%).

%
%
Blooming unfolding imitates flower blooming motion which continuously  unfolds a polyhedron onto a foldable net without collisions~\cite{pak2009cut}. 
Consider a function $u_B: G \rightarrow N$ that maps a goal shape $G$ to a blooming net $N$.
The collision-free motion of $u_B$ implies that the volume swept by unfolding $G$ to $N$ is smaller than other nets. 
Thus, a blooming net reduces the possibility of collision with the laser line of sight and substrate, making it the best option for lasergami.  
Although it is known that all convex shapes have blooming unfoldings~\cite{demaine2011continuous}, 
applying $u_B$ to non-convex shapes, in general, does not create a net that admits continuous blooming \cite{hao2018creat}. 
In fact, as illustrated in Fig.~\ref{fig:pipeline}(\textbf{B}), when $G$ is not convex, $u_B(G)$ is most likely an unfolding containing overlaps, \emph{i.e.,} not a laser-formable net. 
Therefore, 
we propose the idea of a NB net (Alg.~S1) that unfolds $G$ into a net with the provably fewest ``edits" from a blooming unfolding (Thm.~S1-S2).

A motion planner instructs step-by-step where and when the laser should be emitted so the resulting folding motion satisfies all constraints. 
Solving a motion planning problem requires significant computation, but there are several reasons that make planning necessary: 
(A) The nearly-blooming net lowers the chance of collisions during folding but does not eliminate them.  
(B) While some laser occlusions can only be resolved by modifying the nets (Fig.~S6), 
many laser occlusions are avoidable by folding some creases before others under the outside-in folding order (Fig.~\ref{fig:overview}\textbf{(B)}).  
Because of the outside-in folding order, 
one might suspect that the configuration space of a net is highly constrained and low-dimensional.
However, we proved that the planning still requires time complexity exponential to the number of faces in the net (Thm.~S3). 

When a net cannot be fully folded, we observe that many collisions and occlusions can be avoided if we can tolerate that some creases are not fully folded,  
leading to a second key technique for enabling lasergami: a new folding motion planner that can fold a net as close to its goal as possible 
by \emph{unfolding folded creases} as little as possible. 
Fig.~\ref{fig:pipeline}(\textbf{C}) shows, in comparison to the traditional motion planner (MP)~\cite{xl-iros-15}, FP motion planning can fold the net significantly closer to the goal shape (26.25\%	 vs. 97.20\%).

\begin{sloppypar}
The main advantages of the proposed NB unfolding and FP   planner were explored for a diverse set of goal shapes, some of them portrayed in Fig.~\ref{fig:results}. 
All NB nets folded by the FP planner show superior laser formability averaging 96.08\% fold completion, compared with GA nets folded by the FP planner, NB nets folded by the traditional MP, and GA nets folded by the traditional MP, having 75.47\%, 46.87\%, and 13.31\% average fold completions respectively (Fig.~S13). 
The fold completion results for three goal shapes (bunny, whale, and snail) are shown in Fig.~\ref{fig:results}. 
Additionally, we observe that NB nets clip only 71.80\% of substrate area compared to nets created using GA (Fig.~S14). 
\end{sloppypar}

The proposed process is highly flexible, and additional optimization may 
further improve performance~\cite{ma2020metal} by reducing energy, error, or the number of faces that require clipping as shown in Fig.~\ref{fig:pipeline}\textbf{(D)}.   
Significant $1/10$th fabrication energy $E$ was saved for the bunny shape using additional optimization, with only $F=11$ faces requiring clipping compared to the initial $49$ faces.
The FP motion planner may also integrate approximations of the thermal field to enable even faster laser manufacturing~\cite{hao2020planning}. 

\paragraph{Conclusions.} Without the assistance from computational design, human's cognitive capability  is limited to creating simple  lasergami structures.
This work unleashed the potential of lasergami and demonstrated
a diverse set of metallic shapes previously believed to have been unfeasible to be laser-formed.
This contribution was enabled by new computational methods, namely {\em nearly-blooming unfolding} and {\em as-folded-as-possible planning}. 
While the novel proposed methods showed effectiveness 
and allowed for rapid, straightforward, and cost-effective fabrication of 3D shapes using a budget laser cutter,
the structural complexity, fabrication precision, and time efficiency can be further increased using more advanced laser cutters with more degrees of freedom.

\bibliography{scibib,georobotics,geom,origami,masc,ml}

\bibliographystyle{Science}


\section*{Acknowledgments}
\paragraph{Funding:} Y.H.~and J.-M.L. were partially supported by US DOA W911NF1920121. P.J.M.~and E.A.P.H.~gratefully acknowledge the 2020 Henry Samueli School of Engineering Research Grant and UC Irvine startup funds. 
\paragraph{Author contributions: }
Y.H.~and J.-M.L.~contributed to the development and computational implementation of the manufacturing framework and the fabrication of samples. 
H.P.~contributed to the manufacturing of the laser fabrication setup and the bugbot application. 
P.J.M.~and E.A.P.H.~created the models for energy and shape error of the lasergami process and contributed to the  optimization of the determined nets.  
E.A.P.H.~and J-M.L.~designed and supervised the research. 
All authors contributed to the preparation of the manuscript.
\paragraph{Competing interests:} 
The authors declare no conflict of interest. 
\paragraph{Data and materials availability:} 
All results are reproducible with the supplementary materials.

\end{document}